\UseRawInputEncoding
\documentclass[runningheads,hyphens]{llncs}

\usepackage{array}
\newcolumntype{L}[1]{>{\raggedright\arraybackslash}p{#1}}
\newcolumntype{C}[1]{>{\centering\arraybackslash}p{#1}}
\usepackage{graphicx}
\usepackage{caption}
\usepackage{float}
\usepackage{subcaption}
\usepackage[pagebackref,breaklinks,colorlinks,allcolors=blue]{hyperref}

\newcommand{\tb}{\textcolor[RGB]{0, 0, 0}}

\usepackage{makecell}

\usepackage{eccv}

\usepackage{eccvabbrv}

\usepackage{graphicx}
\usepackage{booktabs}

\usepackage[accsupp]{axessibility}  

\usepackage{orcidlink}

\begin{document}

\title{\vspace{-5pt}\tb{Automating Visual Recognition of Leprosy\\in Wild~Chimpanzees\vspace{-15pt}}}

\titlerunning{Leprosy Recognition in Chimpanzees}

\author{Katie I. Murray\inst{1}$^*$\orcidlink{0009-0007-2042-2285} \and
Anna C. Bowland\inst{1}\orcidlink{0009-0007-3905-9041} \and
Marina Ramon\inst{1}\orcidlink{0000-0002-1637-7844} \and
Elena Bersacola\inst{1}\orcidlink{0000-0003-3814-8687} \and
Aissa Regalla\inst{3}\orcidlink{0000-0003-1522-8211} \and
Manmohan D. Sharma\inst{1}\orcidlink{0000-0002-9957-3153} \and
Markus Mueller\inst{1}\orcidlink{0000-0001-7489-6397} \and
Majid Mirmehdi\inst{2}\orcidlink{0000-0002-6478-1403} \and
Dave Hodgson\inst{1}\orcidlink{0000-0003-4220-2076} \and
Kimberley J. Hockings\inst{1}\orcidlink{0000-0002-6187-644X} \and
Tilo Burghardt\inst{2}\orcidlink{0000-0002-8506-012X} \and
Otto Brookes\inst{2}\orcidlink{0000-0001-6865-1844}}

\authorrunning{K.I.~Murray et al.}

\institute{
University of Exeter, Centre for Ecology and Conservation, Penryn, UK
\and
University of Bristol, School of Computer Science,
Bristol, UK 
\and
Wild Chimpanzee Foundation, Leipzig, Germany 
\and
Instituto da Biodiversidade e das \'Areas Protegidas (IBAP), Bissau, Guinea-Bissau
}

\maketitle

\begin{abstract}
    \tb{Leprosy (\textit{Mycobacterium leprae}) has been confirmed in wild western chimpanzees (\textit{Pan troglodytes verus}) in West Africa, presenting as clear and progressive visual symptoms. Manual review of camera-trap footage at landscape scale is infeasible, motivating the need for automated screening. We present the first deep learning pipeline for wildlife leprosy detection and contribute the PanLep300 dataset of 125,670 annotated bounding-box crops across 953 tracks from 303 camera-trap videos with ecologically-motivated splits that withhold whole individuals and camera installations. We benchmark spatial (2D), temporally aggregated (2.5D), and video-based (3D) classification approaches to investigate which approach is best suited to automated leprosy detection in wild apes. We find that simple aggregation of crop-level predictions consistently matches or outperforms both learned temporal models and end-to-end video architectures -- consistent with leprosy's static cutaneous presentation. We further find that performance is suppressed when tracklets contain frames of partially visible individuals -- as commonly occurs at the start and end of a track -- and demonstrate that this can be addressed through targeted construction and aggregation strategies.}
    
\end{abstract}


\section{Introduction}
\label{sec:intro}\vspace{-6pt}

\begin{figure}[t]
\centering
\includegraphics[width=1.02\linewidth, height=180pt]{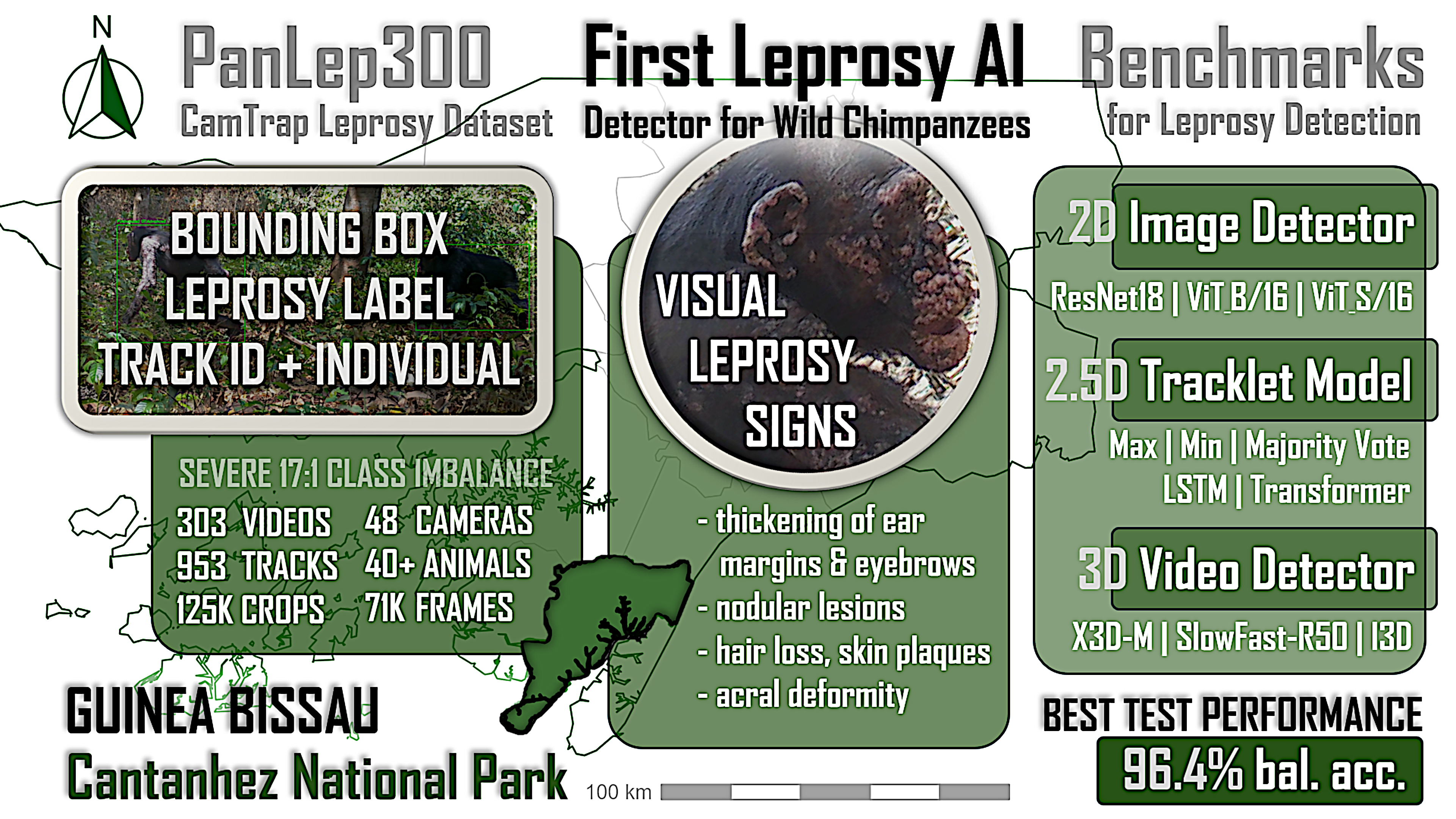}\vspace{-8pt}
\caption{\tb{\small\textbf{Conceptual Overview.} The PanLep300 Dataset comprises 125,670 annotated bounding-box crops across 71,508 frames of wild chimpanzee camera-trap footage with track-level leprosy labels. We benchmark image classifiers, temporal aggregators, and end-to-end video models surfacing actionable findings on the value of explicit localisation, the hazards of background masking, and tracklet boundary dilution.}}
\label{fig:overview}\vspace{-14pt}
\end{figure}

\begin{figure}[t]\vspace{-8pt}
\centering
\includegraphics[width=\linewidth]{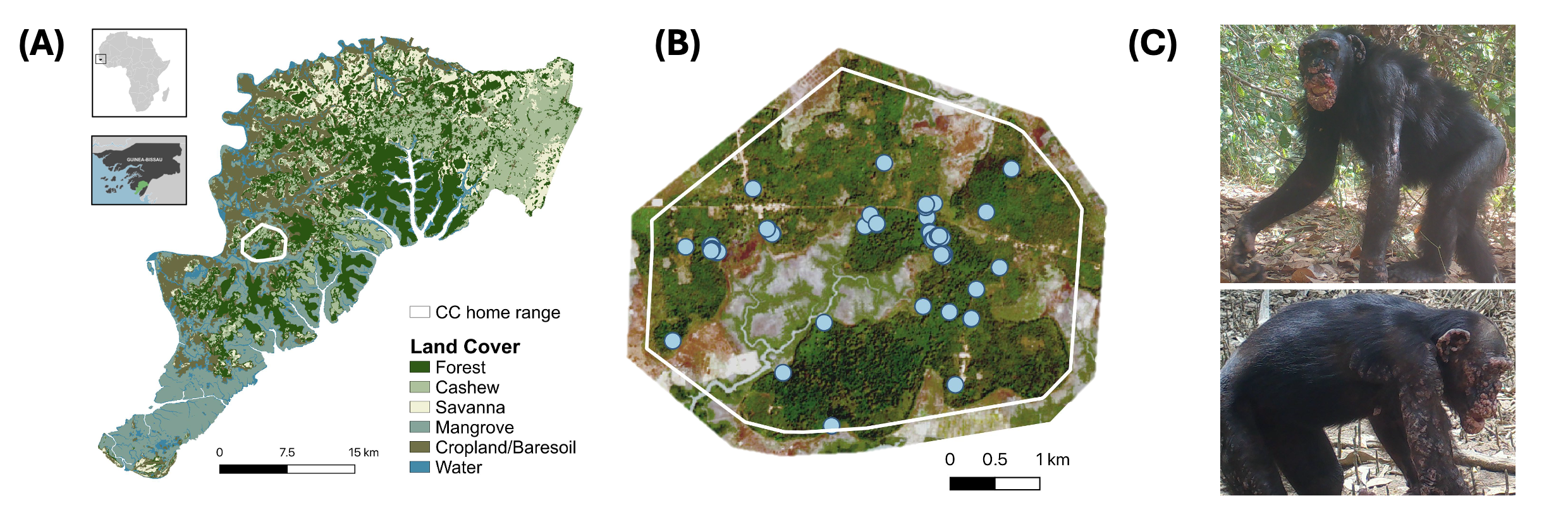}\vspace{-8pt}
\caption{\tb{\small\textbf{Data Collection at Cantanhez
National Park.} \textbf{(A)}~Research area and land cover within CNP~\cite{pereira2022mapping}, with the study community's range at Caiquene-Cadique~(CC) outlined in white. \textbf{(B)}~Camera-trap placements (light blue points) within the community's home range, estimated via 100\% MCP analysis of 1,380 chimpanzee observations (2013--2018); some points overlap due to paired or repeated placements across years. \textbf{(C)}~Examples of chimpanzees exhibiting clearly detectable leprosy signs. Base maps from Natural Earth; Sentinel-2 imagery from Sentinel Hub, Sinergise LTD.}}\vspace{-12pt}
\label{fig:map}
\end{figure}

\tb{\textbf{Ecological Motivation}. The western chimpanzee (\textit{Pan troglodytes verus}) is critically endangered~\cite{iucn_2020_regional,gilardi2025update}, with an estimated 80\% population decline since 1990~\cite{kuhl2017critically}. Infectious disease is a predominant driver of chimpanzee population decline~\cite{iucn_2020_regional,dunay2018pathogen,negrey2019simultaneous}. Chimpanzees' close genetic relatedness to humans renders them highly susceptible to zoonotic pathogens~\cite{calvignac2012wild} -- among them \textit{Mycobacterium leprae}, the causative agent of leprosy, a chronic infectious disease with an incubation period ranging from months to 30 years in humans~\cite{britton2004leprosy,world2021towards}. Leprosy manifests across a spectrum of clinical manifestations that vary in severity according to the host immune response~\cite{ridley1966classification,walker2007leprosy}. At the severe (lepromatous) end, it produces persistent cutaneous signs in chimpanzees~\cite{suzuki2011chimpanzees,hockings2021leprosy} reliably identifiable in camera-trap footage~(see~Fig.~\ref{fig:overview}), including thickened ear margins and eyebrows, skin depigmentation, nodular lesions, hair loss, and hand/foot deformation~\cite{hockings2021leprosy}~(see~Fig.~\hyperref[fig:map]{\ref*{fig:map}-C}). Such signs have been detected across at least six social groups in Cantanhez National Park (CNP), Guinea-Bissau, with molecular confirmation in one~\cite{hockings2021leprosy,bersacola2023action}. We target this severe end of the spectrum, labelling individuals with visible signs as \textit{Advanced} and those without as \textit{None}; the latter denotes the absence of visually identifiable signs, not confirmed absence of infection. Camera-traps represent the only feasible means of systematic landscape-level leprosy monitoring for unhabituated wild chimpanzees, yet deployments generate footage volumes far exceeding expert review capacity, creating an annotation bottleneck~\cite{tuia2022perspectives} that restricts surveillance coverage and delays response.}



\tb{\textbf{State-of-the-art \& Limitations.} Robust methodological progress in automated wildlife monitoring is closely tied to the availability of dedicated, domain-specific datasets. In recent years, a growing number of camera-trap benchmarks have been introduced to support a range of computer vision tasks, including species classification, individual re-identification, behaviour recognition, and multi-animal tracking~\cite{schall2026gorillawatch,gadot2024crop, wasmuht2025sa, brookes2025panaf}. By contrast, automated disease recognition from camera-trap imagery remains limited: only a small number of studies have applied learned models to this task~\cite{muneza2019quantifying,nurccin2024automated,tichon2026monitoring,ringwaldt2026epidemiological}, and no publicly available dataset ready for machine learning~(ML) utilisation exists for the detection of wildlife disease from camera-trap data~\cite{brundage2026generating}. To our knowledge, no prior work has applied automated deep learning (DL) disease detection to wild great apes nor automated DL-based leprosy detection to wildlife in general.}

\tb{\textbf{Contribution}. In response, we present the first automated DL pipeline for leprosy screening in wildlife. Specifically, our contributions are: \textit{\textbf{(1)}}~PanLep300, a wildlife leprosy dataset of 
125,670 annotated bounding-box crops across 71,508 frames, 953 tracks, 303 videos, and 48 camera locations spanning five years (2021--2025) of footage, with per-individual identity and leprosy labels and an ecologically-motivated split design that withholds named individuals and unseen cameras for evaluation; \textit{\textbf{(2)}}~a systematic evaluation of spatial~(2D), temporally aggregated~(2.5D), and video-based~(3D) leprosy classification; and \textit{\textbf{(3)}}~three actionable findings: individual-level localisation substantially outperforms full-frame classification; simple aggregation matches or outperforms more complex architectures, such as end-to-end video models; and partial visibility at track boundaries suppresses leprosy recognition in a predictable and correctable way.}\vspace{-11pt}

\section{Related Work}\vspace{-6pt}

\tb{\textbf{AI for Conservation -- Great Apes}. Wildlife monitoring is fundamental to conservation, yet the scale and complexity of modern survey programmes exceed the capacity of manual analysis~\cite{tuia2022perspectives}. Computer vision and animal biometrics~\cite{kuhl2013animal} have emerged as scalable tools to expedite this analysis, with DL systems for a wide range of camera-trap bio-monitoring tasks~\cite{norouzzadeh2018automatically,schneider2019past, wasmuht2025sa}. Several methods and datasets have been developed specifically for non-human great apes, addressing localisation~\cite{schofield2019chimpanzee, brookes2024panaf20k}, individual identification~\cite{iashin2025self, schall2026gorillawatch, laskowski2023gorillavision}, and behaviour recognition~\cite{brookes2023triple, brookes2025panaf, mueller2026privi, jing2026ethoclip}. None of these works, however, extend to disease detection -- nor the specific case of leprosy -- in unhabituated wild populations.}

\tb{\textbf{AI for Disease Detection in Humans}. Outside the wildlife domain, AI-based disease classification is well-established in human clinical settings~\cite{de2025usage}. Convolutional Neural Networks~(CNNs) detect diabetic retinopathy with sensitivity and specificity surpassing board-certified ophthalmologists~\cite{gulshan2016}, and classify malignant skin lesions, including melanoma, at dermatologist-level accuracy~\cite{esteva2017}.
For leprosy, Baweja~\textit{et al.}~\cite{baweja2023leprosy} achieved 98\% accuracy on controlled dermatological images using a custom CNN (LeprosyNet), while the WHO Skin Neglected Tropical Diseases programme~\cite{deps2026independent} achieved high performance on classical lesions~(84.9\% top-5 sensitivity), declining substantially though for reactional and atypical presentations~(69.6\%), highlighting the challenge of generalising across lesion diversity. These studies demonstrate the feasibility of automated leprosy detection in principle, but our setting introduces additional challenges: camera-trap footage includes partial occlusion, variable subject distance, and cross-camera domain shift, none of which are present in curated clinical datasets.}

\tb{\textbf{AI for Disease Detection in Wildlife}. Camera-traps have long supported manual workflows for wildlife disease surveillance where signs are visually detectable~\cite{barroso2025pixelated}, yet few studies have automated this with artificial intelligence~\cite{muneza2019quantifying,nurccin2024automated}. Nur\c{c}in~\textit{et al.}~\cite{nurccin2024automated} classified devil facial tumour disease using U-Net segmentation with ResNet features on captive Tasmanian devil images. Tichon~\textit{et al.}~\cite{tichon2026monitoring} monitored five diseases over six years in Nubian ibex via camera-traps and generalised additive models, using natural markings for individual-level tracking. Muneza~\textit{et al.}~\cite{muneza2019quantifying} applied the Jenks natural breaks algorithm to quantify giraffe skin disease severity from camera-trap imagery. Most closely related to our work, Ringwaldt~\textit{et al.}~\cite{ringwaldt2026epidemiological} trained a CNN to detect rumpwear -- a visually apparent skin condition -- in common brushtail possums across a three-year, landscape-scale camera-trap network in Tasmania to predict disease prevalence. Brundage~\textit{et al.}~\cite{brundage2026generating} showed that synthetic training images of alopecia can support screening (0.85 AUROC, sim-to-real), while highlighting the need for real, annotated benchmarks. A fundamental barrier across this body of work is data scarcity: no publicly available, ML-ready dataset exists for wildlife health conditions in camera-trap imagery~\cite{brundage2026generating}. This work directly addresses that gap through the release of an annotated dataset PanLep300.}

\section{Dataset}\vspace{-4pt}

\tb{\textbf{Acquisition Context.} The PanLep300 dataset was collected within Cantanhez National Park (CNP)~(see~Fig.~\ref{fig:map}), a 1,067 km\textsuperscript{2} agroforest mosaic~\cite{pereira2022mapping} in the Tombali region of Guinea-Bissau, West Africa (11$^{\circ}$14.2870'N, 15$^{\circ}$02.2810'W). As an IUCN Category V Protected Area situated within the Upper Guinean Forest biodiversity hotspot~\cite{myers2000biodiversity}, CNP supports approximately 25,000 people across more than 200 villages, alongside a minimum of 12 communities of critically endangered western chimpanzees (\textit{Pan troglodytes verus}), unhabituated to researchers, with approximately 35-60 individuals per community~\cite{bersacola2021chimpanzees,iucn_2020_regional,hockings2021leprosy}. The close phylogenetic relationship between humans and chimpanzees, combined with their shared use of forest-mangrove-grassland-agriculture mosaics, creates substantial opportunities for pathogen transmission~\cite{dunay2018pathogen,bersacola2022examining}. Overlap in space and resource use, including wild foods, crops, and water sources, can increase exposure risk, even in the absence of direct contact~\cite{dunay2018pathogen,negrey2019simultaneous}.}

\tb{\textbf{Disease Context.} Leprosy was first detected in CNP chimpanzees through camera-trap analysis in 2015 and subsequently confirmed via molecular analysis of faecal samples~\cite{hockings2021leprosy}. Severe cases of leprosy produce characteristic, persistent cutaneous signs, including nodules, plaques, hypopigmentation, hair loss, and acral deformities such as claw hand, that can be reliably identified in camera-trap imagery~\cite{hockings2021leprosy,suzuki2011chimpanzees}. This principle likely extends to other diseases with a persistent visible phenotype~\cite{barroso2025pixelated}; the pipeline presented here may therefore offer a viable approach for detecting other visually identifiable pathologies in other taxa.}\vspace{-10pt}

\begin{table}[t]
\centering
\caption{\tb{\small\textbf{PanLep300 Dataset Split Statistics.} Non-empty frames are annotated frames containing at least one visible individual. Tracklets are non-overlapping $T{=}16$ frame sequences constructed using the padded variant, duplicating the last frame for clip completion. \% lep.\ denotes the percentage of each unit (videos, frames, crops, tracks, or tracklets) containing at least one individual with \textit{Advanced} leprosy label.}}\vspace{-2pt}
\label{tab:splits}
\scriptsize
\setlength{\tabcolsep}{5pt}
\begin{tabular}{lrrrrrr}
\toprule
Split & \makecell{Videos \\ (\% lep.)} & \makecell{Non-empty frames \\ (\% lep.)} & \makecell{Crops \\ (\% lep.)} & \makecell{Tracks \\ (\% lep.)} & \makecell{Tracklets \\ (\% lep.)} \\
\midrule
Train & 204 (23.0) & 48,384 (8.3) & 80,687 (5.0) & 604 (8.9) & 5,300 (5.3) \\
Val   &  30 (23.3) &  7,327 (13.8) & 12,332 (9.4) &  93 (10.8) &   808 (9.3) \\
Test  &  69 (21.7) & 15,797 (13.2) & 32,651 (6.7) & 256 (6.6)  & 2,165 (6.7) \\
\textbf{Total} & \textbf{303} & \textbf{71,508} & \textbf{125,670} & \textbf{953} & \textbf{8,273} \\
\bottomrule
\end{tabular}\vspace{-7pt}
\end{table}

\subsection{Dataset Overview}\vspace{-2pt}

\tb{\textbf{Key Statistics.} The corpus comprises 125,670 bounding-box crops extracted from 71,508 annotated frames across 953 tracks in 303 videos, covering 48 distinct cameras. Tab.~\ref{tab:splits} summarises split-level statistics; the overall positive fraction is 0.054 at crop level (a 17:1 negative-to-positive ratio; see also Fig.~\hyperref[fig:track_distribution_IDonly]{\ref*{fig:track_distribution_IDonly}-B} for track-level), reflecting the low disease prevalence encountered in the field. Bounding-box sizes vary considerably as a consequence of the uncontrolled field setting: subjects are filmed at distances ranging from close-range trail crossings to distant forest-edge sightings, and the three camera models differ in focal length and sensor resolution. The most data-rich camera position~(TREC1) contributes 19,342 crops across 92 tracks; the camera with the highest track count~(CAM1) records 107 distinct individual appearances. Mean crops per track are 132.7~(median:~74), reflecting considerable variation from brief transits to extended sequences; the mean number of annotated tracks per camera position is 19.7.}\vspace{-10pt}

\subsection{Camera Deployment}\vspace{-2pt}

\tb{\textbf{Field Data Acquisition.} Camera-trap data was contributed to us by the Cantanhez Chimpanzee Project~(CCP) [\small{\url{https://cantanhezchimpanzeeproject.com}}\normalsize] who work in close  partnership with the Instituto da Biodiversidade e das \'Areas Protegidas~(IBAP) [\small{\url{https://ibapgbissau.org}}\normalsize]. All chimpanzees in CNP are unhabituated to human presence, making camera-traps the only feasible means of systematic landscape-level disease monitoring. Although some cameras were deployed specifically for disease monitoring, the resulting observations remained opportunistic, with individuals  filmed under variable conditions of camera placement, viewing angle, and proximity that influenced symptom visibility.}

\tb{\textbf{Recording Specifications.} Three commercial camera-trap models were deployed across 48 recording locations over five years (2021--2025), including 3~arboreal installations and 4 paired camera stations providing near-$360^{\circ}$ coverage of transit paths. Cameras record at 30 or 60~fps across varying resolutions and sensor configurations, introducing variation in input format -- in addition to natural variation in background foliage, illumination, distance to subject, and body orientation -- before any pathology-related variation is considered. Each camera is identified by a prefix code in the dataset filename (e.g.\ \texttt{CC\_CAM01}, \texttt{CC\_MANG}).}\vspace{-8pt}

\subsection{Ground Truth Data Annotation}\vspace{-2pt}

\begin{figure}[t]\vspace{-8pt}
\centering
\includegraphics[width=\linewidth]{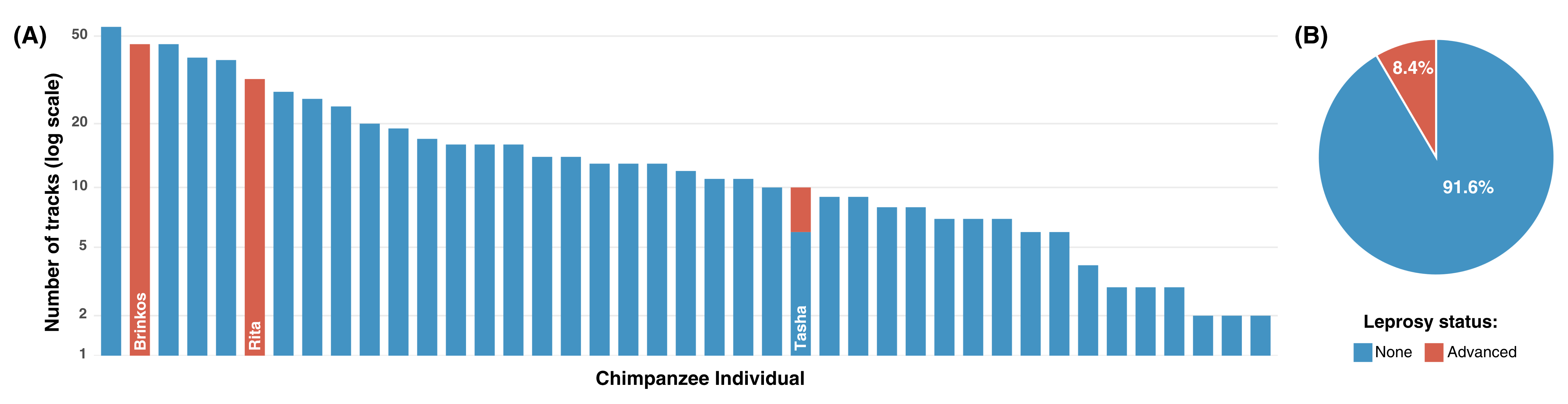}\vspace{-8pt}
\caption{\tb{\small\textbf{Track Distribution across Chimpanzee Individuals.} Distribution and leprosy symptom status across PanLep300. \textbf{(A)}~Number of tracks per named individual~(41 identity classes), ordered by total track count and coloured by symptom status: \textit{Advanced} (red; visible leprosy signs) and \textit{None} (blue; no observable symptoms), plotted on a log scale. Three non-individual classes -- \textit{chimpanzee} (115 tracks), \textit{Infant} (199 tracks), and \textit{Sub-adult} (27 tracks), all with leprosy attribute \textit{None} -- are excluded for clarity. \textbf{(B)}~Overall track distribution across all 44 identity classes~(41~uniquely named, 3~infant/sub-adult/other), showing the pronounced class imbalance between tracks of individuals with signs of
leprosy (\textit{Advanced}) and without (\textit{None}).}}\vspace{-12pt}
\label{fig:track_distribution_IDonly}
\end{figure}

\tb{\textbf{Labelling Regime.} Bounding-box annotations with track IDs were initially machine-generated using MegaDetectorV6~\cite{beery2019efficient,hernandez2024pytorchwildlife} with SORT tracking~\cite{bewley2016simple}, then reviewed and corrected in CVAT~\cite{sekachev2020opencv} by researchers with extensive field experience of the CNP chimpanzee population and expertise in identifying leprosy symptoms. For every visible chimpanzee in each sampled frame, annotators verified the accuracy of the individual's bounding box and track ID, and assigned a track-level class label for individual ID (44 chimpanzee name classes;~see~Fig.~\ref{fig:track_distribution_IDonly}) and a binary attribute label, \texttt{leprosy}~$\in$~\{\textit{Advanced}, \textit{None}\}, which remains fixed for the duration of the track. \textit{Advanced} (label~1) indicates that the individual displays at least one clinically recognisable sign of leprosy: depigmented skin patches, nodular lesions on the face or limbs, digit loss, limb deformity, or postural asymmetry~\cite{hockings2021leprosy}. \textit{None} (label~0) denotes the absence of any visible symptoms. Of the 44 identity classes, two represent age categories rather than named individuals (\textit{infant} and \textit{sub-adult}), and one (\textit{chimpanzee}) was assigned to individuals that could not be reliably identified due to distance or occlusion. The remaining 41 classes each correspond to a uniquely identified individual (see~Fig.~\hyperref[fig:track_distribution_IDonly]{\ref*{fig:track_distribution_IDonly}-A}). Note that the dataset includes a named individual, \emph{Tasha}, whose disease status (and therefore label) changes across the recording period~(see~Fig.~\hyperref[fig:track_distribution_IDonly]{\ref*{fig:track_distribution_IDonly}-A}).}

\tb{\textbf{Sampling Regime.} Raw video is downsampled to a universal rate of 10~fps (every 6th frame at 60~fps; every 3rd frame at 30~fps), preserving smooth inter-frame motion while reducing a typical one-minute clip from approximately 3,600 to 600 frames. Bounding boxes are drawn as tight crops around each individual's visible body and interpolated between manually placed keyframes; per-frame visibility flags~(\texttt{outside}, \texttt{occluded}) are recorded for every box. Crops flagged \texttt{outside=1}, where the individual has left the frame or is not visible, are excluded from all splits.}

\vspace{-7pt}
\subsection{Ecology-motivated Split Design}\vspace{-3pt}
\label{sec:splits}

\tb{\textbf{Dataset Splits.} Naive random splitting would inflate test metrics, since camera and individual-specific cues are strong confounders. We instead apply a hierarchical, video-level strategy governed by three rules, applied in priority order:}\vspace{-4pt}

\begin{enumerate}
    \item \tb{\textbf{Disjoint Individuals}: Tracks depicting four named individuals -- \emph{Tasha}, \emph{NA\_1} , \emph{NA\_2} , and \emph{Elf} -- are assigned exclusively to the test set. \emph{Tasha}'s inclusion prevents identity-based shortcuts, since her tracks span both disease states. \emph{NA\_1} , \emph{NA\_2} , and \emph{Elf} contribute negative test examples from individuals withheld entirely from training.}
    \item \tb{\textbf{Disjoint Camera Locations}: All tracks from two camera locations -- prefixes \texttt{CC\_MANG} (mango orchard habitat, a visually distinct open-canopy environment) and \texttt{CC\_CAM07} (a novel interior forest scene) -- are reserved exclusively for test, ensuring evaluation on unseen hardware/environment footage.}
    \item \tb{\textbf{Stratified Video-Level Allocation}: Remaining videos are stratified by positive-track presence, with all tracks from a given video assigned to a single partition to prevent within-video crop leakage across splits.\vspace{-4pt}}
\end{enumerate}

\noindent \tb{The resulting test set therefore contains individuals the model cannot have memorised, cameras whose visual characteristics were absent during training, and a named individual whose disease trajectory explicitly rules out identity as a proxy for class membership.}



\vspace{-9pt}
\section{PanLep300 Benchmarks}\vspace{-4pt}

\tb{We benchmark three method classes for PanLep300: \textbf{(1)}~\emph{2D} image-based classifiers on full frames and bounding-box crops, isolating the effect of individual localisation; \textbf{(2)}~\emph{2.5D} sequence models operating on frozen spatial features, isolating the contribution of temporal context; and \textbf{(3)}~\emph{3D} end-to-end spatio-temporal video models, capturing both jointly.}
\vspace{-10pt}
\subsection{Setup}\vspace{-4pt}

\tb{\textbf{Training Environment.} All models are trained with cross-entropy loss, AdamW (weight decay~0.05), RandAugment (2~ops, magnitude~9), and cosine-with-warmup scheduling (10\% warmup, min LR factor~0.01) for 100~epochs. Image and video models are initialised with ImageNet~\cite{deng2009imagenet} and Kinetics~\cite{carreira2017quo} pre-trained weights, respectively. All images undergo short-side scaling to the target resolution~(224 or 336~px). To maintain aspect ratio when scaling, images are padded with ImageNet-mean grey. Video models are trained and tested on tracklets. Each full length track is partitioned into non-overlapping clips of~$T{=}16$ frames. We evaluate two tracklet construction approaches: the \textit{padded} variant, where the final incomplete clip is completed by duplicating its last frame, and the \textit{trimmed} variant, where remainder frames are removed symmetrically from both ends of each track: for remainder $R = N \bmod T$, trim $\lfloor R/2 \rfloor$ from the start and $\lceil R/2 \rceil$ from the end (e.g., a 53-frame track at $T{=}16$ yields 3 clean tracklets instead of 4 padded ones). To combat class imbalance, we follow the protocol proposed by Kang \textit{et al.}~\cite{kang2019decoupling}, referred to as stage B results here. Specifically, we freeze the backbone at the best Stage~A checkpoint, re-initialise the head, and retrain for 50~epochs on a Weighted Random Sampling (WRS) balanced stream.}

\tb{\textbf{Evaluation Strategy}. We structure our evaluation around increasing temporal scope. We first evaluate image-level classification, comparing full-frame and crop-level inputs (Sec.~\ref{2d}). We then move to a spatio-temporal evaluation, considering two units of analysis: \textit{tracklets}, fixed-length subclips sampled from within a track (Sec.~\ref{track}), and \textit{full tracks}, the complete variable-length sequence of crops belonging to a single individual (Sec.~\ref{tracklet}). Results are reported separately at both granularities, since aggregation behaviour -- and consequently model ranking -- differs between the two.}

\subsection{Image-level Evaluation (2D) -- Frames vs. Crops}\vspace{-4pt}
\label{2d}

\begin{figure}[t]\vspace{-8pt}
  \centering
  \subfloat[\tb{Frame-level Incorrect \& Crop-level Correct}]{\includegraphics[width=0.48\textwidth]{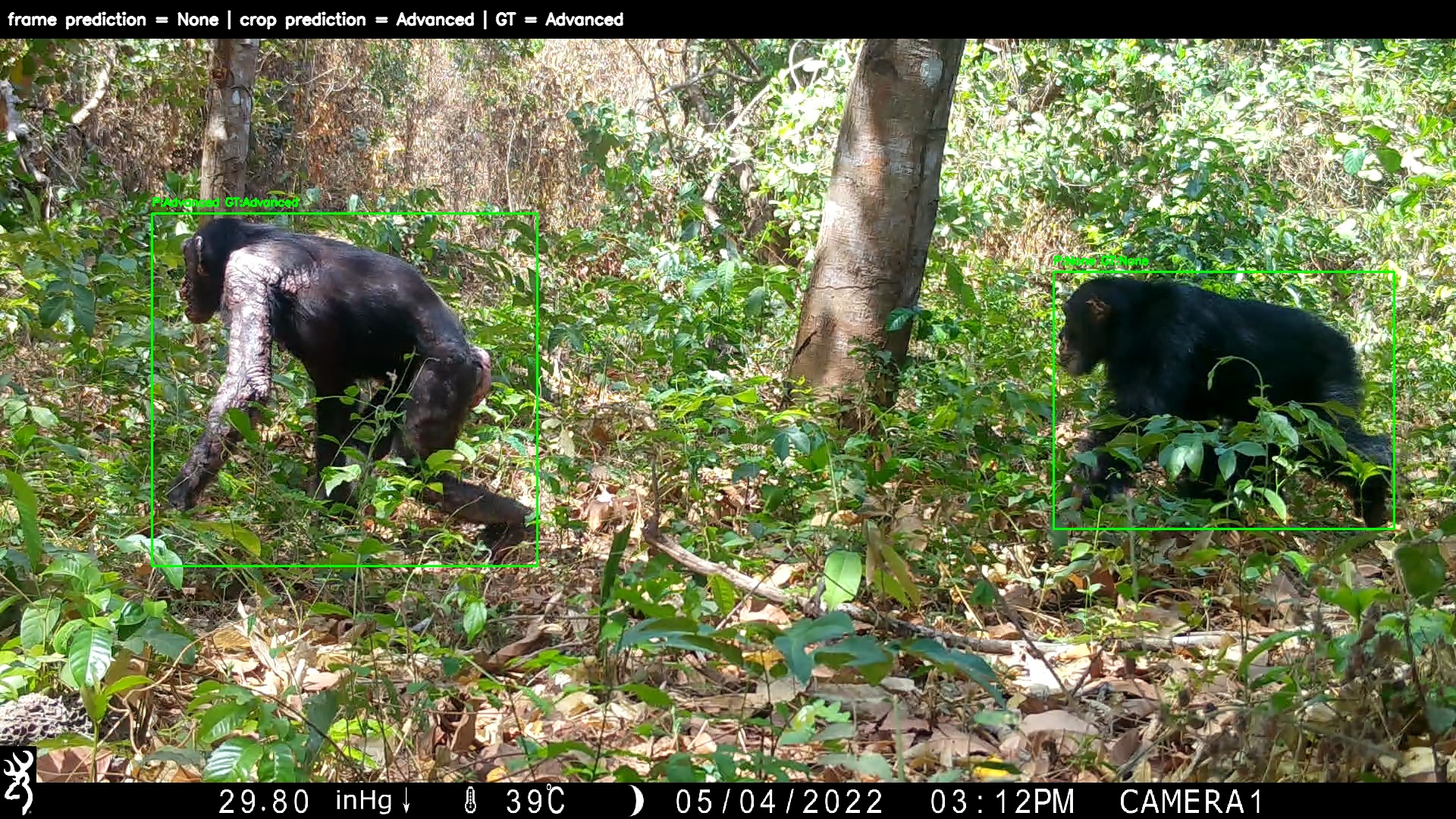}\label{fig:f1}}
  \hfill
  \subfloat[\tb{Frame-level Correct \& Crop-level Incorrect}]{\includegraphics[width=0.48\textwidth]{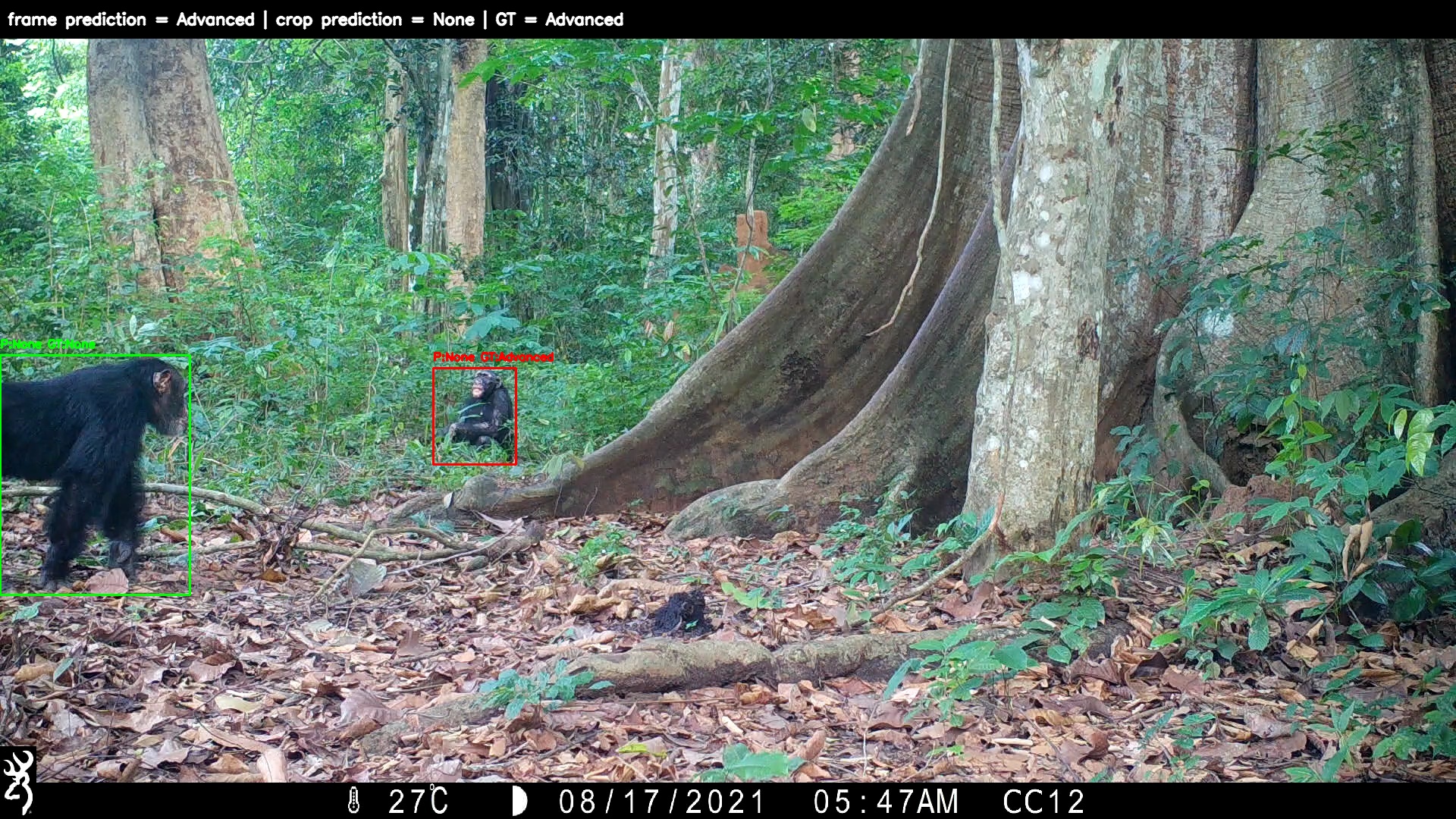}\label{fig:f2}}\vspace{-6pt}
  \caption{\tb{\textbf{Qualitative Comparison of Frame and Crop-level Classification.} Test frame examples where full-frame classification and crop-level classification disagree at frame-level. Bounding boxes indicate ground-truth region localisation, coloured green (correctly localised) or red (missed). The header on each frame shows full-frame model prediction, crop-aggregated frame-level prediction, and ground-truth label.}}\vspace{-16pt}\label{fig:f}
\end{figure}

\tb{\textbf{Experimental Overview}. We first evaluate the extent to which image classifiers can predict the presence of leprosy in camera-trap imagery. We implement convolutional and attention-based architectures~--~ResNet18, ViT-S/16, and ViT-B/16~--~under two scenarios: full-frame and crop-level classification~(see Fig.~\ref{fig:f}). In the full-frame setting, a frame is labelled positive if any individual present exhibits leprosy (an any-positive rule). In the crop-level setting, classification is performed at the level of individual bounding-box crops. To enable direct comparison, crop-level predictions are aggregated to the frame level via the same any-positive rule. Finally, for background removal experiments, the Segment Anything Model 3~(SAM3)~\cite{carion2025sam,wasmuht2025sa} is applied to all frames using the prompt `chimpanzee'. Resulting background pixels are replaced with a neutral grey to prevent models learning correlated background information.}\vspace{-2pt}

\begin{table}[!hb]\vspace{-12pt}
\centering
\caption{\tb{\textbf{Crop-Level vs. Full-Frame Classification at Increasing Resolution}. Balanced accuracy, F1$_{+}$, precision, and recall are compared across ResNet-18, ViT-S/16, and ViT-B/16 under full-frame classification at 224$\times$224 and 896$\times$896, and crop classification aggregated to frame level. In general, increasing full-frame resolution yields modest gains for ResNet-18 and ViT-S/16, but crop-level classification outperforms full-frame classification at any resolution tested, across all architectures.}}\vspace{-6pt}
\label{tab:full-frame_vs_crop}
\scriptsize
\setlength{\tabcolsep}{3.5pt}
\begin{tabular}{L{2cm} C{1.5cm} C{1.7cm} C{1.2cm} C{1.2cm} C{1.2cm}}
\toprule
Model & Bal.\ Acc. & Top-1 Acc. & F1+ & Prec.+ & Rec.+ \\
\midrule
\midrule
\multicolumn{6}{l}{\itshape Full-frame classification (224$\times$224)} \\
ResNet-18  & 0.662 & 0.920 & 0.469 & 0.794 & 0.333 \\
ViT-S/16             & 0.757 & 0.927 & 0.656 & 0.873 & 0.526 \\
ViT-B/16             & 0.721 & 0.915 & 0.587 & 0.817 & 0.458 \\
\midrule
\multicolumn{6}{l}{\itshape Full-frame classification (896$\times$896)} \\
ResNet-18  & 0.741 & 0.918 & 0.617 & 0.807 & 0.500 \\
ViT-S/16   & 0.766 & 0.902 & 0.612 & 0.646 & 0.582 \\
ViT-B/16   & 0.694 & 0.908 & 0.537 & 0.800 & 0.404 \\
\midrule
\multicolumn{6}{l}{\itshape Crop classification -- frame-level evaluation} \\
ResNet-18  & \textbf{0.866} & 0.914 & 0.712 & 0.641 & \textbf{0.801} \\
ViT-S/16   & 0.843 & \textbf{0.952} & \textbf{0.793} & \textbf{0.920} & 0.696 \\
ViT-B/16   & 0.828 & 0.942 & 0.754 & 0.857 & 0.672 \\
\bottomrule
\end{tabular}
\end{table}

\tb{\textbf{Cropping Yields More Precise Classification than Resolution Scaling Alone}. As shown in Tab.~\ref{tab:full-frame_vs_crop}, full-frame classification at 224$\times$224 achieves a best balanced accuracy of 0.757 (ViT-S/16), but with consistently poor recall across all architectures (0.333--0.526), suggesting that without explicit localisation, models frequently miss positive individuals. Increasing input resolution to 896$\times$896 partially closes this gap -- balanced accuracy improves for ResNet-18 (0.662 $\to$ 0.741) and ViT-S/16 (0.757 $\to$ 0.766), though ViT-B/16 instead degrades slightly (0.721 $\to$ 0.694) -- but even the best higher-resolution result (0.766) still falls short of every crop-level result in the table. Aggregating crop-level predictions to the frame level via an any-positive rule yields substantially larger gains than resolution scaling alone, with ResNet-18 achieving the highest overall balanced accuracy (0.866) driven by strong recall (0.801), and ViT-S/16 achieving the best F1$_{+}$ (0.793) at higher precision (0.920). This indicates that the benefit of cropping is not simply a function of effective spatial resolution of the lesion site, but of removing irrelevant scene content and other individuals from the input entirely; explicit localisation is therefore a critical step that resolution scaling alone cannot replace, justifying bounding-box annotation.}

\tb{\textbf{ResNets and ViTs Maximise Recall or Precision, Respectively}. Tab.~\ref{tab:crop-level_vs_masks} summarises crop-level performance across architectures. ResNet-18 Stage~A clearly achieves the highest balanced accuracy (0.878) driven by strong recall (0.789), making it the most sensitive detector at the crop level and the natural choice where missing a positive case carries the greatest cost. The ViT models achieve lower balanced accuracy but substantially higher precision (0.918 for both ViT-S/16 and ViT-B/16 vs.\ 0.622 for ResNet18), at the cost of recall (0.679 and 0.657, respectively). The trade-off is consistent across both ViT scales, with ViT-S/16 offering marginal gains in balanced accuracy over ViT-B/16 (0.837 vs.\ 0.827) despite its smaller parameter count, suggesting that model scale yields diminishing returns at the crop level.}

\tb{\textbf{Background Masking Degrades Balanced Accuracy.} As shown in Tab.~\ref{tab:crop-level_vs_masks}, SAM-based background masking reduces balanced accuracy across all architectures: $-$8.9\% for ResNet18, $-$4.0\% for ViT-S/16, and $-$1.0\% for ViT-B/16. }\vspace{-2pt}

\begin{table}[!ht]\vspace{-12pt}
\centering
\caption{\tb{\textbf{Crop-Level Classification vs.\ Background-Masked Classification.} Balanced accuracy, F1$_{+}$, precision, and recall are compared across ResNet-18, ViT-S/16, and ViT-B/16 under standard crop-level classification and SAM-based background masking. In general, masking reduces balanced accuracy across all architectures, with the largest drop for ResNet-18 and the smallest for ViT-B/16, while precision improves for ResNet-18 under masking despite the overall decline.}}\vspace{-6pt}
\label{tab:crop-level_vs_masks}
\scriptsize
\setlength{\tabcolsep}{3.5pt}
\begin{tabular}{L{2cm} C{1.5cm} C{1.7cm} C{1.2cm} C{1.2cm} C{1.2cm}}
\toprule
Model & Bal.\ Acc. & Top-1 Acc. & F1+ & Prec.+ & Rec.+ \\
\midrule
\midrule
\multicolumn{6}{l}{\itshape Crop classification -- crop-level evaluation} \\
ResNet-18  & \textbf{0.878} & 0.954 & 0.696 & 0.622 & \textbf{0.789} \\
ViT-S/16   & 0.837 & \textbf{0.974} & \textbf{0.781} & \textbf{0.918} & 0.679 \\
ViT-B/16   & 0.827 & 0.970 & 0.766 & 0.918 & 0.657 \\
\midrule
\multicolumn{6}{l}{\itshape Mask classification} \\
ResNet-18  & 0.789 & 0.963 & 0.683 & 0.816 & 0.587 \\
ViT-S/16   & 0.797 & 0.967 & 0.707 & 0.858 & 0.601 \\
ViT-B/16   & 0.817 & 0.970 & 0.739 & 0.874 & 0.640 \\
\bottomrule
\end{tabular}
\end{table}

\noindent\tb{However, the effect is not uniform across metrics. For ResNet18, masking increases precision (0.816 vs.\ 0.622) and Top-1 accuracy (0.963 vs.\ 0.954) whilst substantially reducing recall (0.587 vs.\ 0.789). The ViT models exhibit a similar but less pronounced pattern. The consistent reduction in balanced accuracy indicates that background context contributes a positive discriminative signal, whether through habitat cues correlated with the disease -- which~is unlikely -- or because SAM mask inaccuracies remove some lesion-bearing regions.}
\vspace{-13pt}
\subsection{Tracklet-level Evaluation (2.5D)}\vspace{-6pt}
\label{tracklet}

\tb{\textbf{Experimental Overview}. Having established that crop-level classification is advantageous, we next investigate whether aggregating predictions across a sequence of crops from the same individual yields further gains. We evaluate three approaches to this problem: \textbf{\textit{(1)}} first, the best-performing crop model, selected by balanced accuracy, is frozen and applied independently to each crop in a track; the resulting per-crop class probabilities are aggregated to a single track-level prediction via mean probability ($\bar{p} = \frac{1}{n}\sum p_i$, positive if $\geq 0.5$), max probability ($p^* = \max_i p_i$), and majority vote. We refer to these as heuristic or hand-coded aggregation. \textbf{\textit{(2)}} Secondly, we train lightweight aggregators -- a mean-pool MLP, a bi-LSTM, and a Transformer encoder -- on frozen crop features extracted from the same backbone, allowing the model to learn a temporal weighting rather than relying on a fixed statistic. \textbf{\textit{(3)}} Finally, for comparison against these crop-based pipelines, we implement three video-native architectures -- I3D~\cite{carreira2017quo}, X3D~\cite{feichtenhofer2020x3d}, and SlowFast-R50~\cite{feichtenhofer2019slowfast} -- each initialised with Kinetics-400-pretrained weights~\cite{carreira2017quo}, which operate directly on raw frame sequences rather than aggregated crop predictions. The same set of clips~($T{=}16$ or $T{=}32$) is used throughout.}

\begin{table}[!ht]\vspace{-13pt}
\centering
\caption{\tb{\textbf{Tracklet-level Aggregation: Heuristic vs.\ Trained vs.\ End-to-End.} Balanced accuracy, F1$_{+}$, and recall are compared across three aggregation strategies -- heuristic over Stage~A crop probabilities, trained temporal aggregators on frozen crop features, and end-to-end 3D video models -- at two clip lengths ($T{=}16$, $T{=}32$). Hand-coded aggregation achieves the best balanced accuracy throughout.}}\vspace{-6pt}
\label{tab:tracklet_level}
\scriptsize
\setlength{\tabcolsep}{3pt}
\begin{tabular}{L{2.8cm} C{1.4cm} C{1.1cm} C{1.1cm} C{0.15cm} C{1.4cm} C{1.1cm} C{1.1cm}}
\toprule
 & \multicolumn{3}{c}{$T = 16$} & & \multicolumn{3}{c}{$T = 32$} \\
\cmidrule(lr){2-4} \cmidrule(lr){6-8}
Model & Bal.\ Acc. & F1+ & Rec.+ & & Bal.\ Acc. & F1+ & Rec.+ \\
\midrule
\multicolumn{8}{l}{\itshape Heuristic aggregation (2D ResNet-18 Stage~A crops)} \\
Majority vote        & \textbf{0.875} & 0.775 & \textbf{0.766} & & \textbf{0.879} & 0.790 & 0.771 \\
Mean probability     & 0.874 & 0.763 & 0.766 & & 0.879 & \textbf{0.797} & 0.771 \\
Max probability      & 0.845 & 0.430 & 0.836 & & 0.825 & 0.351 & \textbf{0.869} \\
\midrule
\multicolumn{8}{l}{\itshape Trained temporal aggregators (frozen 2D ResNet-18 Stage~A backbone)} \\
Transformer          & 0.842 & 0.789 & 0.688 & & 0.836 & 0.796 & 0.672 \\
Mean-pool MLP        & 0.842 & 0.709 & 0.703 & & 0.833 & 0.759 & 0.672 \\
Bi-LSTM              & 0.819 & 0.722 & 0.648 & & 0.834 & 0.774 & 0.672 \\
\midrule
\multicolumn{8}{l}{\itshape 3D video models, Stage~A (end-to-end on raw frames)} \\
X3D-M                & 0.849 & \textbf{0.793} & 0.703 & & 0.813 & 0.630 & 0.656 \\
SlowFast-R50         & 0.838 & 0.791 & 0.680 & & 0.824 & 0.748 & 0.656 \\
I3D                  & 0.802 & 0.726 & 0.609 & & 0.841 & 0.771 & 0.689 \\
\bottomrule
\end{tabular}
\end{table}

\tb{\textbf{Simple Aggregation Proves Superior}. As shown in Tab.~\ref{tab:tracklet_level}, hand-coded aggregation of Stage~A per-crop probabilities (i.e., heuristic aggregation) remains the strongest approach overall, with majority vote and mean probability jointly achieving the best balanced accuracy at both clip lengths (0.874--0.879). Neither the trained temporal aggregators nor the 3D video models close this gap -- all six fall short of the best hand-coded results at both clip lengths, with X3D-M the closest competitor at $T{=}16$ (0.849) 
and I3D the closest at $T{=}32$ (0.841) on balanced accuracy. 
The best of the 3D video models holds a small edge over the best of the trained aggregators at both clip lengths (0.849 vs.\ 0.842 at $T=16$; 0.841 vs.\ 0.836 at $T=32$). In general, however, the video models do not largely or consistently outperform the trained aggregators despite operating directly on raw frame sequences rather than fixed crop features, and in several cases score lower still. X3D-M's balanced accuracy falls from 0.849 to 0.813 between $T{=}16$ and $T{=}32$, the steepest degradation of any method. While results are not fully comparable across the image-based (see~Sec.~\ref{2d}) and spatio-temporal pipelines given their differing input representations, the absence of any clear benefit from end-to-end spatio-temporal modelling is notable at this scale. If motion or temporal dynamics carried a substantial diagnostic signal, the 3D models would be expected to exploit this advantage over frame-independent aggregation. Their failure to do so is consistent with leprosy's clinical presentation as a set of static cutaneous and postural signs rather than a behavioural or motion-dependent symptom, suggesting that the appearance of crops carries the bulk of the signal.}

\begin{table}[!ht]\vspace{-8pt}
\centering
\caption{\tb{\textbf{Track-level Aggregation Results.} Using 256 test tracks (17~pos., 239~neg.), heuristic methods aggregate Stage~A per-crop probabilities over all crops per track. Trained temporal aggregators and 3D video models evaluate all non-overlapping $T{=}16$ padded clips per track, aggregating via mean-softmax.}}\vspace{-7pt}
\label{tab:track_level}
\scriptsize
\setlength{\tabcolsep}{4pt}
\begin{tabular}{L{3cm} C{1.4cm} C{1.8cm} C{1.1cm} C{1.1cm} C{1.1cm}}
\toprule
Model & Bal.\ Acc. & Top-1 Acc. & F1+ & Prec.+ & Rec.+ \\
\midrule
\multicolumn{6}{l}{\itshape Heuristic aggregation (all crops per track)} \\
Mean probability     & \textbf{0.941} & 0.992 & \textbf{0.938} & \textbf{1.000} & 0.882 \\
Majority vote        & 0.941 & 0.992 & 0.938 & 1.000 & 0.882 \\
Max probability      & 0.918 & 0.848 & 0.466 & 0.304 & \textbf{1.000} \\
\midrule

\multicolumn{6}{l}{\itshape Trained aggregators, padded clips + mean-softmax, $T{=}16$} \\
Transformer          & 0.853 & 0.977 & 0.828 & 1.000 & 0.706 \\
Mean-pool MLP        & 0.851 & 0.973 & 0.800 & 0.923 & 0.706 \\
Bi-LSTM              & 0.794 & 0.969 & 0.741 & 1.000 & 0.588 \\
\midrule
\multicolumn{6}{l}{\itshape 3D video models, padded clips + mean-softmax, $T{=}16$} \\
X3D-M                & 0.853 & \textbf{0.980} & 0.828 & 1.000 & 0.706 \\
SlowFast-R50         & 0.838 & 0.953 & 0.667 & 0.632 & 0.706 \\
I3D                  & 0.824 & 0.977 & 0.786 & 1.000 & 0.647 \\
\bottomrule
\end{tabular}
\end{table}

\vspace{-10pt}
\subsection{Full Track-level Evaluation (3D)}\vspace{-5pt}
\label{track}

\tb{\textbf{Experimental Overview}. The preceding sections evaluate models on fixed-length tracklets, a useful control, but a departure from how disease screening would operate in practice. A camera-trap deployment yields full tracks of variable length. In this section, we therefore apply the previously trained models to entire tracks, evaluating each at the unit on which it would ultimately be deployed. This means classifying every crop in the full-length track and pooling the resulting probabilities. Note this reduces the total number of test samples significantly~--~from 1,923 trimmed- or 2,165 padded- tracklets to just 256 tracks.}

\begin{figure}[!hb]
\centering
\includegraphics[width=1.045\linewidth]{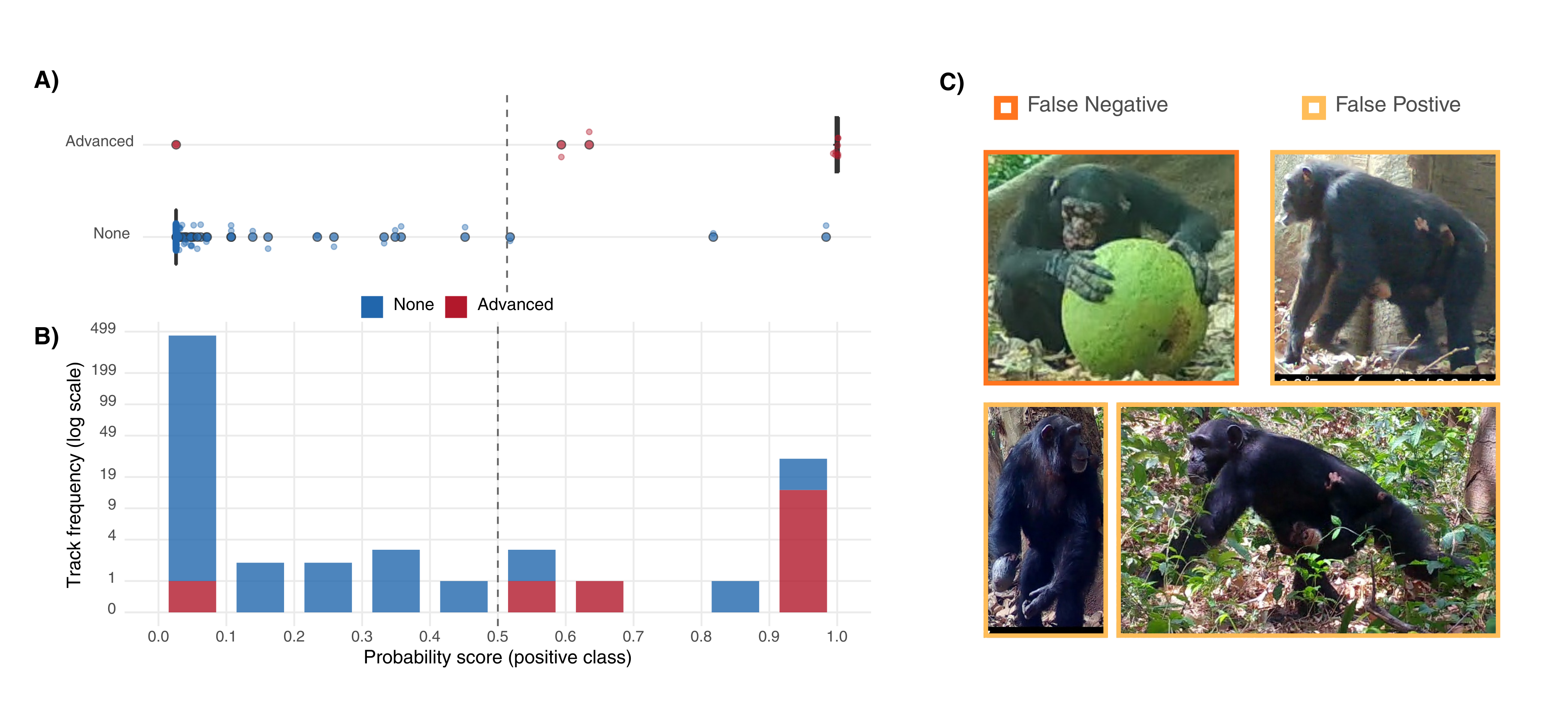}\vspace{-16pt}
\caption{\tb{\textbf{Predicted Leprosy-likelihood Distribution for X3D-M.} Data shown for Stage-A-only, trimmed $T{=}16$, top-25\% aggregation on the 256-track test set. \textbf{A)}~Box-and-whisker plots for None (blue) and Advanced (red) leprosy in tracks; dashed line marks the 0.5 threshold. \textbf{B)}~Log-scale histogram across 10 probability bins. The distribution is strongly bimodal: 226/239 negative tracks score below 0.05 and 14/17 positive tracks score above 0.95 with 16/17~TP, 3~FP, 1~FN and bal.\ acc.\ at 0.964. \textbf{C)}~An example crop from each of the incorrectly predicted tracks: 1 FN and 3 FP.}}
\end{figure}

\begin{figure*}[!h]
\includegraphics[width=\textwidth,height=137pt]{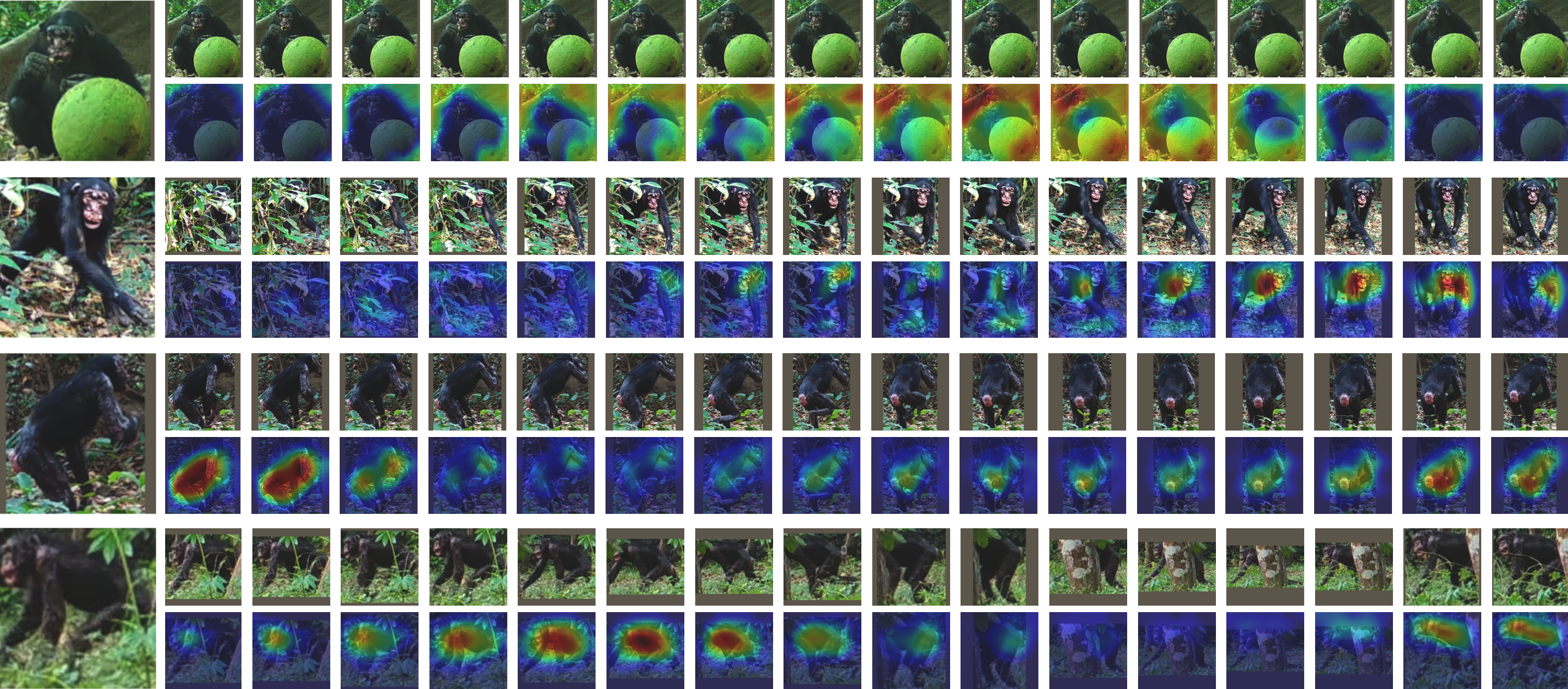}\vspace{-4pt}
\caption{\tb{\textbf{Grad-CAM Visualisations of Leprosy Classifier for X3D-M.} Shown are four 16 frame skims Stage-A-only, trimmed $T{=}16$, at top-25\% aggregation. \textbf{Top~Skim}~shows false negative with attention collapse to the foreground treculia fruit (\textit{Treculia africana}) -- all models fail on this track. \textbf{2nd} shows true positive with near-zero activation during occlusion. \textbf{3rd} shows true positive with tightly localised attention hotspots on the upper body concentrated around clear symptom areas. \textbf{Bottom} shows true positive with activation dissipation as the animal is behind a tree.}}
\label{fig:short}\vspace{-16pt}
\end{figure*}

\tb{\textbf{Heuristic Aggregation Remains Superior at Full Track-level.} As shown in Tab.~\ref{tab:track_level}, the pattern observed at the tracklet level persists, and is even more pronounced, when evaluating on full-length tracks. Mean probability and majority vote jointly achieve the best balanced accuracy (0.941) with perfect precision (1.000), correctly identifying 15 of 17 positive tracks without a single false alarm. Max probability again trades this precision for perfect recall (1.000) at severe cost to precision (0.304), confirming that a single high-scoring crop is sufficient to flag an entire track positive under this rule -- a pattern consistent with its behaviour at the tracklet level. The trained temporal aggregators and 3D video models all fall short of hand-coded aggregation, and the gap widens relative to the tracklet-level results (see Tab.~\ref{tab:tracklet_level}). The Transformer is the strongest of the trained aggregators (0.853) but still trails mean probability by nearly 9\%, while the bi-LSTM and mean-pool MLP fall further behind (0.794 and 0.851, respectively). The 3D video models perform comparably to or worse than the trained aggregators despite operating directly on raw frames, with X3D-M the strongest (0.853)~(see also Fig.~\ref{fig:short}) and I3D the weakest (0.824); notably, SlowFast-R50 is the only 3D model with imperfect precision (0.632) more prone to false alarms.}\vspace{-14pt}

\begin{table}[!h]
\centering
\caption{\tb{\textbf{Track-Level Evaluation Across Aggregation Strategies.} Balanced accuracy, F1$_{+}$, and Recall$_{+}$ are compared side-by-side across trained temporal aggregators for Stage-A-only $T{=}16$ on 256 test tracks.}}\vspace{-7pt}
\label{tab:track_level_side_by_side}
\scriptsize
\setlength{\tabcolsep}{3pt}
\begin{tabular}{l ccc c ccc}
\toprule
 & \multicolumn{3}{c}{\textbf{Padded}} & & \multicolumn{3}{c}{\textbf{Trimmed}} \\
\cmidrule{2-4} \cmidrule{6-8}
Model & Bal. Acc. & F1+ & Rec.+ & & Bal. Acc. & F1+ & Rec.+ \\

\midrule

\multicolumn{8}{l}{\itshape \textbf{Trained temporal aggregators}} \\
~\textit{Mean-softmax Strategy} & & & & & & & \\
~~~Transformer    & 0.853 & 0.828 & 0.706 & & 0.853 & 0.828 & 0.706 \\
~~~Bi-LSTM        & 0.794 & 0.741 & 0.588 & & 0.794 & 0.741 & 0.588 \\
~~~Mean-pool MLP  & 0.851 & 0.800 & 0.706 & & 0.824 & 0.786 & 0.647 \\

~\textit{Top-$\alpha$ strategy, all $\alpha=25$\%} & & & & & & & \\
~~~Transformer & 0.901 & 0.778 & 0.824 & & 0.941 & 0.938 & 0.882 \\
~~~Bi-LSTM & 0.937 & 0.882 & 0.882 & & 0.939 & 0.909 & 0.882 \\
~~~Mean-pool MLP & 0.885 & 0.509 & 0.882 & & 0.950 & 0.744 & 0.941 \\

~\textit{Top-$\alpha$ Strategy$^\ast$} & & & & & & & \\
~~~Transformer (p:25\%, t:25\%)   & 0.901 & 0.778 & 0.824 & & 0.941 & 0.938 & 0.882 \\
~~~Bi-LSTM (p:33\%, t:50\%)    & 0.939 & 0.909 & 0.882 & & 0.941 & 0.938 & 0.882 \\
~~~Mean-pool MLP (p:33\%, t:25\%)   & 0.904 & 0.600 & 0.882 & & 0.950 & 0.744 & 0.941 \\

\midrule
\multicolumn{8}{l}{\itshape \textbf{3D video models, Stage~A}} \\
~\textit{Mean-softmax strategy} & & & & & & & \\
~~~X3D-M          & 0.853 & 0.828 & 0.706 & & \textbf{0.912} & 0.903 & 0.824 \\
~~~SlowFast-R50   & 0.838 & 0.667 & 0.706 & & 0.882 & 0.867 & 0.765 \\
~~~I3D            & 0.824 & 0.786 & 0.647 & & 0.824 & 0.786 & 0.647 \\

~\textit{Top-$\alpha$ strategy$^\ast$} & & & & & & & \\
~~~X3D-M (p:33\%, t:25\%)        & 0.872 & 0.743 & 0.765 & & \textbf{0.964} & 0.889 & 0.941 \\
~~~SlowFast-R50 (p:50\%, t:25\%) & \textbf{0.931} & 0.615 & 0.941 & & 0.939 & 0.909 & 0.882 \\
~~~I3D (p:33\%, t:25\%)  & 0.920 & 0.714 & 0.882 & & 0.878 & 0.813 & 0.765 \\
\bottomrule
\multicolumn{7}{l}{\scriptsize \textit{$^\ast$ $\alpha$ chosen per architecture to maximise balanced accuracy}} \\
\end{tabular}
\label{tab:3d_detail}
\end{table}\vspace{-8pt}

\tb{\textbf{Boundary Tracklets Dilute Confident Interior Predictions}. Per-tracklet probabilities reveal a systematic gap between boundary and interior chunks. Of the 17 positive test tracks, 10 score uniformly high (>0.95) throughout; the remaining 7 show clear boundary dilution, with boundary tracklets (first and last) averaging 0.291 against 0.595 for middle tracklets. This reflects a common pattern in the footage: diagnostic signal is often concentrated mid-track, when an individual is briefly unoccluded, while boundary frames capture partial views as it enters or exits the frame. Mean-softmax aggregation, which weights every chunk equally regardless of diagnostic content, lets these near-zero boundary probabilities pull the track mean below threshold even when interior chunks are confidently positive. On one representative track, for instance, SlowFast scores two interior chunks at $p>0.99$ but three boundary chunks at $p\approx0$, yielding a track mean of 0.431 -- a confident positive diluted to a false negative. Three of the seven affected tracks fail this way: middle tracklets clear the decision threshold but are dragged below it by their boundary counterparts. We address this with two independent strategies: \textbf{\textit{(1)}} \textit{trimmed construction}, which removes remainder frames symmetrically from both ends of each track instead of padding~--~for remainder $R = N \bmod T$, $\lfloor R/2 \rfloor$ frames are trimmed from the start and $\lceil R/2 \rceil$ from the end, yielding 3 clean tracklets instead of 4 padded ones for a 53-frame track at $T{=}16$, at a cost of 10--16\% fewer tracklets overall; and \textbf{\textit{(2)}} \textit{top-$\alpha$\% mean aggregation}, which averages only the top $\lceil \frac{\alpha}{100} \times N \rceil$ chunks by probability ($\alpha \in \{25, 33, 50, 67, 75\}$) rather than all, filtering out uninformative boundary chunks while retaining the noise-reducing benefit of averaging. The two strategies act independently, at construction and aggregation stage respectively.}

\tb{\textbf{Trimmed Construction and Top-$\alpha$\% Aggregation Can Help Independently and Additively.} As shown in Tab.~\ref{tab:3d_detail}, trimmed construction alone improves balanced accuracy over padded tracklets for X3D-M (0.853 $\to$ 0.912) and SlowFast-R50 (0.838 $\to$ 0.882) under the same mean-softmax aggregation, confirming that padding artefacts are a genuine contributor to dilution rather than a symptom that aggregation alone can resolve. Replacing mean-softmax with top-$\alpha$\% aggregation on top of trimmed tracklets yields a further, larger gain for X3D-M and SlowFast-R50, reaching the best results in the table: 0.964 for X3D-M (top-25\%) and 0.939 for SlowFast-R50 (top-25\%). I3D is the exception throughout -- trimming produces no change at all (0.824 in both rows).}\vspace{-8pt}

\section{Conclusions (Discussion and Limitations)}\vspace{-4pt}
\tb{We present the first automated deep learning pipeline for leprosy screening in any wildlife and release a dataset (PanLep300) of 125,670 annotated crops across 953 tracks, with an ecologically-motivated split to test generalisation. Our systematic evaluation across spatial, temporally aggregated, and video-based architectures confirms maximal effectiveness of simple aggregation of crop-level probabilities, a finding consistent with leprosy's static cutaneous presentation rather than any behavioural or motion-dependent symptom. It offers a strong first detector for leprosy in wild apes and mirrors observations in human clinical AI, where single frames often suffice for dermatological conditions with visible phenotypes~\cite{esteva2017,baweja2023leprosy}.}

\tb{Several limitations remain. The test set contains only 17 positive tracks, making close comparisons statistically fragile; larger-scale validation across additional sites and seasons is needed before the pipeline could operate as a stand-alone screening tool. The pipeline currently requires ground-truth bounding boxes, as well as integration with an automated detector and tracker for future field deployment. The study leaves disentanglement of genuine ecological correlation from confounding effects of camera placement open for future work, particularly as the pipeline is extended to new sites (and potentially new species).}

\tb{Our results nevertheless demonstrate that automated AI disease screening from camera-trap footage alone can be feasible and effective. Infectious disease is an increasingly recognised driver of wildlife population decline~\cite{iucn_2020_regional,cunningham2017one}, yet the data volumes generated by modern camera-trap networks render manual review challenging at landscape scale~\cite{tuia2022perspectives,barroso2025pixelated}. The approach presented here is generalisable beyond leprosy and chimpanzees to any clearly visually identifiable pathology in any species accessible to camera-trapping. Thus, we hope that this work, and the dataset and findings it provides, will support continued monitoring of leprosy in this critically endangered chimpanzee population and inform the design of similar systems for other wildlife disease surveillance efforts.}

\bibliographystyle{splncs04}
\bibliography{main}
\end{document}